\documentclass[sigconf]{acmart-me}
\usepackage{booktabs} % For formal tables
\usepackage{url}
\usepackage{color}
\usepackage{enumitem}
\usepackage{bm}
\setcopyright{none}
\acmDOI{}

\acmISBN{}

\acmConference[*This work was completed during Jennifer's research internship with Google. \\ MediaEval'18]{Multimedia Evaluation Workshop}{29-31 October 2018}{Sophia Antipolis, France} 
\acmYear{}
\copyrightyear{}

\acmPrice{}

\begin{document}
\title{GLA in MediaEval 2018 Emotional Impact of Movies Task}
%\titlenote{Produces the permission block, and
%  copyright information}
%\subtitle{Extended Abstract}
%\subtitlenote{The full version of the author's guide is available as
 % \texttt{acmart.pdf} document}

%%If you want to use multi-column authors that's fine. Comment out the next lines, and then uncomment below.
\author{Jennifer J. Sun\textsuperscript{1,}*, Ting Liu\textsuperscript{2},Gautam Prasad\textsuperscript{2}}
\affiliation{\textsuperscript{1}Caltech, \textsuperscript{2}Google Research}
\email{jjsun@caltech.edu, {liuti,gautamprasad}@google.com}

%%\authornote{The secretary disavows any knowledge of this author's actions.}
%\affiliation{Institute for Clarity in Documentation, Ohio, USA}
%\email{webmaster@marysville-ohio.com}
%
%\author{Lars Th{\o}rv{\"a}ld}
%%\authornote{This author is the
%%  one who did all the really hard work.}
%\affiliation{The Th{\o}rv{\"a}ld Group, Iceland}
%\email{larst@affiliation.org}
%
%\author{Lawrence P. Leipuner}
%\affiliation{Brookhaven Labs, France}
%\email{lleipuner@researchlabs.org}
%
%\author{Sean Fogarty}
%\affiliation{A Research Institute, Germany}
%\email{fogartys@amesres.org}
%\affiliation{Palmer Research Laboratories, Texas, USA}
%\affiliation{The Th{\o}rv{\"a}ld Group, Iceland}
%\email{jsmith@affiliation.org}

\begin{abstract}
The visual and audio information from movies can evoke a variety of emotions in viewers. Towards a better understanding of viewer impact, we present our methods for the MediaEval 2018 Emotional Impact of Movies Task to predict the expected valence and arousal continuously in movies. This task, using the LIRIS-ACCEDE dataset, enables researchers to compare different approaches for predicting viewer impact from movies. Our approach leverages image, audio, and face based features computed using pre-trained neural networks. These features were computed over time and modeled using a gated recurrent unit (GRU) based network followed by a mixture of experts model to compute multiclass predictions. We smoothed these predictions using a Butterworth filter for our final result. Our method enabled us to achieve top performance in three evaluation metrics in the MediaEval 2018 task.
\end{abstract}

%\begin{CCSXML}
%<ccs2012>
%  <concept_id>10010520.10010553.10010562</concept_id>
%  <concept_desc>Computer systems organization~Embedded systems</concept_desc>
%  <concept_significance>500</concept_significance>
%  <concept_id>10010520.10010575.10010755</concept_id>
%  <concept_desc>Computer systems organization~Redundancy</concept_desc>
%  <concept_significance>300</concept_significance>
%  <concept_id>10010520.10010553.10010554</concept_id>
%  <concept_desc>Computer systems organization~Robotics</concept_desc>
%  <concept_significance>100</concept_significance>
%  <concept_id>10003033.10003083.10003095</concept_id>
%  <concept_desc>Networks~Network reliability</concept_desc>
%  <concept_significance>100</concept_significance>
%</ccs2012>  
%\end{CCSXML}
%
%\ccsdesc[500]{Computer systems organization~Embedded systems}
%% We no longer use \terms command
%%\terms{Theory}
%
%\keywords{ACM proceedings, \LaTeX, text tagging}

%% Used in some conference proceedings e.g. sigplan and sigchi
%   \caption{This is a teaser}

\maketitle

\section{Introduction}
\label{sec:intro}
Movies can cause viewers to experience a range of emotions, from sadness to relief to happiness. Viewers can feel the impact of movies, but it is difficult to predict this impact automatically. The ability to automatically predict movie evoked emotions is helpful for a variety of use cases \cite{ref7} in video understanding.    

The Emotional Impact of Movies Task \cite{ref7}, part of the MediaEval 2018 benchmark, provides participants with a common dataset for predicting the expected emotional impact from videos. We focused on the first subtask in the challenge: predicting the expected valence and arousal continuously (every second) in movies. The dataset provided by the task is the LIRIS-ACCEDE dataset \cite{ref3, ref4}, which is annotated with self-reported valence and arousal every second from multiple annotators. Since deep neural networks, such as Inception \cite{ref19}, have millions of trainable parameters, the competition data may be too limited to train these networks from random initialization. Therefore, we used networks that were pre-trained on larger datasets, such as ImageNet, to extract features from the LIRIS-ACCEDE dataset. The extracted features were used to train our temporal and regression models.

This task is recurring with multiple submissions every year \cite{ref11, ref14}. Our method's novelty lies in the unique set of features we extracted including image, audio, and face features (capitalizing on transfer learning) along with our model setup, which comprises of a GRU combined with a mixture of experts.

\section{Approach}
\label{sec:approach}
We approached the valence and arousal prediction as a multivariate regression problem. Our objective is to minimize the multi-label sigmoid cross-entropy loss and this could allow the model to use potential relationships between the two dimensions for regression. We first used pre-trained networks to extract features. To model the temporal aspects of the data, the methods we evaluated included long short-term memory (LSTM), gated recurrent unit (GRU) and temporal convolutional network (TCN). Multiple modalities were fused with late fusion, and valence along with arousal was predicted jointly. Our method is implemented using TensorFlow and we used the Adam optimizer \cite{ref12} in all our experiments.

\subsection{Feature Extraction}
We extracted image, audio and face features from each frame of the movies. Our image features (Inception-Image) were from the Inception network \cite{ref19} pre-trained on ImageNet \cite{ref16}. We extracted audio features using AudioSet \cite{ref9}, which is a VGG-inspired model pre-trained on YouTube-8M \cite{ref1}. For the face features (Inception-Face), we focused on the two largest faces in each frame and used an Inception based architecture trained on faces \cite{ref17}. Since the movies were human-focused, faces were found in most of the scenes. We compared these features with those used in last years competition that included image features computed using VGG16 \cite{ref18} and audio features computed using openSMILE \cite{ref8}. All our features were extracted at one frame per second.

\subsection{Temporal Models}
To model the temporal dynamics of the emotion in the videos, we used recurrent neural networks. In particular, we used LSTMs \cite{ref10} and GRUs \cite{ref6} as part of our modeling pipeline in a sequence-to-one setup with sequence length of 10, 30 or 60 seconds. The self-reported emotions likely depend on past scenes in movies, so temporal modeling is important for this task. In addition, we evaluated TCNs because of their promising performance in sequence modeling \cite{ref2} and action segmentation \cite{ref13}. Specifically, we trained an encoder-decoder TCN using sequences of extracted features to obtain a sequence of valence and arousal predictions.

\subsection{Regression Models}
The input from each modality (image, audio, or face) is fed into separate recurrent models. The output we use from each recurrent model is its hidden state which contains information on previous data seen by the model. We use the hidden state
corresponding to the final timestamp in the input sequence. The state vectors for each modality are concatenated into a single vector and fed into a context gate \cite{ref15}. Multimodal fusion occurs at this stage as we use the learned model to fuse the representation of each modality from the RNN. The output of the context gate is then fed into a mixture-ofexperts model with another context gate to obtain the final emotion predictions. We use logistic regression experts with a softmax gating network.

To prevent overfitting, we regularized our models using L2 regularization, dropout and batch normalization. Finally, a low pass filter is applied on the predictions to smooth the prediction outputs. In LIRIS-ACCEDE, the measured emotion data vary smoothly in time but our regression outputs contain high frequency signals. To smooth our outputs, we tested weighted moving average filters and low-pass filters (Butterworth filter \cite{ref5}) as implemented in SciPy.

\section{Results and Analysis}

We optimized the hyperparameters of our models to have the best performance on the validation set, which consists of 13 movies from the development set. We then trained our models on the entire development set to run inference on the test set. Our setup used a batch size of 512. Through evaluating our recurrent models, we found that Inception-Image+AudioSet features had better performance in terms of MSE and PCC compared to VGG16+openSMILE features. In some cases, the recurrent model would predict near the mean for both valence and arousal while using VGG16+openSMILE. This may be because the features did not have enough information for the models to discriminate between different values of valence and arousal. We also found a significant increase in performance when we added the Inception-Face features, which may point to salient information captured in connection with the expected emotions in the videos.

The sequence-to-one recurrent models worked best with longer input sequences of 60 seconds versus those of 10 or 30 seconds. This observation may be because the evoked emotion is affected by longer lasting scenes. Different input sequence length may work better for videos that are faster or slower paced. Our recurrent models also performed better on the validation set than the TCN and the GRU models had similar performance to the LSTMs. We used GRUs for our implementation because GRUs are computationally simpler than LSTMs. Since we have a small dataset, we wanted to reduce model complexity to prevent underfitting. Our temporal model architecture ranged from 32 to 256 units and 1 to 2 layers, optimized for each of the modalities. 

For post processing, the low-pass Butterworth filter worked better than the moving average filter. This result is likely because the Butterworth filter is designed to have a frequency response as flat as possible (with no ripples) in the pass-band. Fluctuations of the magnitude response within the passband may decrease the accuracy of our regression output.

In Table~\ref{tab:freq} we list the performance of our best models that were submitted to the task. Each of the 5 runs is defined as follows where we used Inception-Image, AudioSet, and Inception-Face as the features and a GRU with mixture-of experts for regression.

\begin{enumerate}
    \item No dropout or batch normalization.
    \item Regularized with dropout and batch normalization. Trained on approximately 70\% of the data.
    \item Regularized with dropout and batch normalization.
    \item Regularized with dropout and batch normalization, different initialization and epoch.
    \item Average over all runs.
\end{enumerate}

We see that creating an ensemble from our models by averaging over the runs has the lowest MSE (Run 5). This is likely because by averaging, we decrease the variance of predictions and thus overall, the mean is closer to the ground truth labels. While averaging improves MSE, it does not improve correlation. Our model with the best correlation is from Run 4. In the results of the MediaEval 2018 Emotional Impact of Movies task \cite{ref7}, we achieved top Valence and Arousal MSE, and top Arousal PCC. 

We note that using batch normalization during inference increases the variance of our predictions. This is because we are using the batch statistics instead of the population statistics from the train set to normalize the batches. Our validation results (with repeated runs) as well as test results show that using batch normalization in this way improves predictions for valence, but not as much for arousal. This is most likely because the statistics of the test set for valence is
different from train set while the test set statistics for arousal may be closer to the train set statistics. One explanation could be the small size of the dataset so that the statistics of the train set does not generalize well to the test set.

\begin{large}
\begin{table}
  \caption{Performance of our five models on the test set in the MediaEval 2018 Emotional Impact of Movies task.}
  \label{tab:freq} 
  \begin{tabular}{c c c c c}
    \toprule
    & Valence & Valence & Arousal & Arousal \\
    & MSE & PCC & MSE & PCC \\
    \midrule
    Run 1 & $ 0.1193$ & $ 0.1175$ & $0.1384$ & $0.2911$\\
    Run 2 & $0.0945$ & $0.1376$ & $0.1479$ & $0.1957$ \\
    Run 3 & $0.1133$ & $0.1883$ & $0.1778$ & $0.2773$ \\
    Run 4 & $0.1073$ & $\bm{0.2779}$ & $0.1396$ & $\bm{0.3513}$\\
    Run 5 & $\bm{0.0837}$ & $0.1786$ & $\bm{0.1334}$ & $0.3358$ \\
  \bottomrule
\end{tabular}
\end{table}
\end{large}

\section{Conclusions}
We found that precomputed features modeling image, audio, and face in concert with GRUs provided the optimal performance in predicting the expected valence and arousal in movies for this task. Other precomputed features may be explored for further improvements in performance.

Based on our test set metrics, ensemble methods such as bagging could be useful for this task. We found some evidence that recurrent models performed better than TCN. However since we only evaluated the encoder-decoder TCN, more investigation will be necessary for a broader conclusion.

The pre-computed features we used to model image, audio, and face information showed better performance when compared with the VGG16+openSMILE baseline. A future direction could be to train the network in an end-to-end manner to better capture the frame level data, with the caveat that we may need a much larger training dataset.

\begin{acks}
We are grateful to the MediaEval competition organizers for providing data and support for this project. We would like to thank Alan Cowen, Florian Schroff, Brendan Jou and Hartwig Adam from Google for helpful discussions. 
\end{acks}

\pagebreak

\bibliographystyle{ACM-Reference-Format}
\def\bibfont{\small} % comment this line for a smaller fontsize
\bibliography{sample-me18} 

\end{document}